\documentclass[sigconf]{acmart}
\usepackage{multirow}
\usepackage{diagbox}
\usepackage{xcolor}
\usepackage{colortbl}
\usepackage{enumitem}
\usepackage{graphicx}
\definecolor{aliceblue}{rgb}{0.94, 0.97, 1.0}
\DeclareUnicodeCharacter{2217}{\ensuremath{-}}
\AtBeginDocument{%
  }

\setcopyright{acmlicensed}
\copyrightyear{2024}
\acmYear{2024}
\acmDOI{10.1145/3627673.3679926}

\acmConference[CIKM '24]{the 33rd
ACM International Conference on Information and Knowledge
Management}{October 21--25, 2024}{Boise, ID, USA}
\acmISBN{979-8-4007-0436-9/24/10}

\begin{document}

\title{FashionLOGO: Prompting Multimodal Large Language Models for Fashion Logo Embeddings}


\author{Zhen Wang}
\authornote{Both authors contributed equally to this research.}
\affiliation{%
  \institution{ByteDance Inc.}
  \city{Beijing}
  \country{China}
}
\email{wz1234@buaa.edu.cn}

\author{Da Li}
\authornotemark[1]
\affiliation{%
  \institution{ByteDance Inc.}
  \city{Beijing}
  \country{China}}
\email{lida.0214@bytedance.com}

\author{Yulin Su}
\authornotemark[1]
\affiliation{%
  \institution{ByteDance Inc.}
  \city{Shanghai}
  \country{China}
}
\email{yulinshh@sjtu.edu.cn}

\author{Min Yang}
\affiliation{%
  \institution{ByteDance Inc.}
  \city{Beijing}
  \country{China}}
\email{yangminbupt@outlook.com}

\author{Minghui Qiu†}
\affiliation{%
  \institution{ByteDance Inc.}
  \city{Shanghai}
  \country{China}}
\email{minghuiqiu@gmail.com}

\author{Walton Wang}
\affiliation{%
  \institution{ByteDance Inc.}
  \city{Shanghai}
  \country{China}}
\email{walton.wang@bytedance.com}


\renewcommand{\shortauthors}{Zhen Wang∗ et al.}

\begin{abstract}
Logo embedding models convert the product logos in images into vectors, enabling their utilization for logo recognition and detection within e-commerce platforms. This facilitates the enforcement of intellectual property rights and enhances product search capabilities. However, current methods treat logo embedding as a purely visual problem. A noteworthy issue is that visual models capture features more than logos. Instead, we view this as a multimodal task, using text as auxiliary information to facilitate the visual model's understanding of the logo. The emerging Multimodal Large Language Models (MLLMs) have demonstrated remarkable capabilities in both visual and textual understanding. Inspired by this, we propose an approach, \textbf{FashionLOGO}, to explore how to prompt MLLMs to generate appropriate text for product images, which can help visual models achieve better logo embeddings. We adopt a cross-attention transformer block that enables visual embedding to automatically learn supplementary knowledge from textual embedding. Our extensive experiments on real-world datasets prove that FashionLOGO is capable of generating generic and robust logo embeddings, achieving state-of-the-art performance in all benchmarks. 
\end{abstract}




\maketitle

\section{Introduction}
Logos are critical in e-commerce for intellectual property protection, brand identity, and product recommendation~\cite{deeplogo2017,fashionbrands}. To accomplish the above tasks, a visual model is needed in which the various logos are embedded as unique vectors. Existing solutions view logo embedding as a purely visual understanding task. Logo embedding models~\cite{hittingdeeplogo,logonet2015, logo:surey} are usually obtained by using convolutional neural networks (CNN) or transformer-based networks as backbone. Unlike natural images, logos are intentionally designed to make a visual impact~\cite{Multi-LabelLogoRecognition}. This means that a logo that occupies a small area will play an equally or even more important role in understanding the image than the remainder of the same picture. But logo embedding models obtained by the above paradigm do not pay enough attention to the logo in the image, because of the small area the logo takes up in the product image. To overcome this difficulty, We try to enhance the model's attention to the logo areas by combining information from other modalities like text.

Recently, the remarkable progress in multimodal large language models (MLLMs) has demonstrated outstanding image comprehension abilities. Notable examples of such models include Valley~\cite{luo2023valleyvideoassistantlarge}, BLIP-2~\cite{DBLP:journals/corr/abs-2301-12597}, InstructBLIP~\cite{DBLP:journals/corr/abs-2305-06500}, Otter \citep{DBLP:journals/corr/abs-2305-03726}, Mini-GPT4~\citep{DBLP:journals/corr/abs-2304-10592} and LLaVA~\citep{DBLP:journals/corr/abs-2304-08485}. To this end, we propose a novel framework called FashionLOGO that leverages the text generated from MLLMs to enhance the original visual representation for logo embedding. Specifically, we first utilize LLaVA~\cite{DBLP:journals/corr/abs-2304-08485} to generate supplementary text based on different designed prompts for all product images. During the training phase, we use the cross-attention method to improve the visual representation of the logo with the help of information from the text. 

Extensive experimental results have conclusively demonstrated the effectiveness of the FashionLOGO, the FashionLOGO shows significant improvements in image retrieval performance compared to previous state-of-the-art methods, showcasing its practical applications in e-commerce, and other fashion-oriented domains. These findings validate the innovative design of the FashionLOGO architecture and highlight its potential to advance the field of fashion-focused visual understanding and retrieval.

\section{Related Work}
\subsection{Logo Recognition}
Logo recognition is a fundamental task in computer vision with a wide range of applications such as brand monitoring, and content-based retrieval. Early approaches relied on hand-crafted features and geometric invariants to capture logo characteristics~\cite{romberg2011scalable,wei2009trademark,zhu2009logo,doermann1993logo,schietse2007practice,sahbi2012context,csurka2004visual}. With the advent of deep learning, convolutional neural networks (CNNs) have become popular for logo recognition~\cite{yu2019cascaded,huang2015vehicle,logodet3k,lou2019veri,hittingdeeplogo,logonet2015,deeplogo2017}. 
Transformer-based methods~\cite{parmar2018image,DBLP:conf/iclr/DosovitskiyB0WZ21} have also shown promising results in logo recognition tasks. 

Recent advancements have focused on developing novel techniques and frameworks to enhance logo embeddings. A deep metric learning \cite{lou2019veri} is proposed for fine-grained logo recognition. The Adversarial Attention Representations(AAR) framework \cite{hu2021makeup216} uses adversarial attention to attend separately to logos and backgrounds. 
An image-text multi-modal pre-training method is introduced in \cite{hubenthal2023image} to improve text sensitivity and adopted the ProxyNCAHN++ loss function to incorporate class-specific hard negative cases. Although these approaches have shown promising results, they focus on improving the discrimination of visual features while neglecting knowledge from other modalities such as text.

\subsection{MLLMs}
As LLMs ~\cite{DBLP:conf/nips/Ouyang0JAWMZASR22,DBLP:journals/corr/abs-2204-02311,DBLP:journals/corr/abs-2302-13971,alpaca,DBLP:journals/corr/abs-2302-13971,DBLP:journals/corr/abs-2210-02414,DBLP:journals/corr/abs-2211-05100,vicuna2023} have demonstrated strong general capabilities in linguistic tasks, enabling LLMs to understand multimodal content has been increasingly studied. Existing methods can be divided into two technical routes, one is to employ LLMs as schedulers to schedule the existing multimodal models, and the other is to train a multimodal model based on LLMs.

After receiving user instructions and the functions of foundation models, the former treats the LLMs as a controller to call corresponding foundation models step by step and integrates the output content of each model to generate final results~\cite{DBLP:journals/corr/abs-2303-04671,DBLP:journals/corr/abs-2303-17580,DBLP:journals/corr/abs-2303-11381}.
For example, HuggingGPT~\cite{DBLP:journals/corr/abs-2303-17580} utilize the ChatGPT~\cite{DBLP:conf/nips/Ouyang0JAWMZASR22} to select appropriate models in Hugging Face\footnote{https://huggingface.co/models} according to their function description and summarizes their execution results. 
The latter equips LLMs with auxiliary modules to help them understand multimodal contents through end-to-end training ~\cite{DBLP:journals/corr/abs-2305-06355,DBLP:journals/corr/abs-2304-10592,damonlpsg2023videollama,DBLP:journals/corr/abs-2304-08485,su2023pandagpt,DBLP:journals/corr/abs-2305-06500}.
For instance, LLaVA \cite{DBLP:journals/corr/abs-2304-08485}, MiniGPT-4 \cite{DBLP:journals/corr/abs-2304-10592}, and Valley~\cite{luo2023valleyvideoassistantlarge} connected LLaMA \cite{DBLP:journals/corr/abs-2302-13971} with a visual encoder through a projection module, endowing it the ability to understand images. Video-LLaMA ~\cite{damonlpsg2023videollama} empowered LLaMA ~\cite{DBLP:journals/corr/abs-2302-13971} with visual and audio information via Q-Former to endow it with video-grounded conversation ability.

\section{Methodology}

In this section, we present our proposed model and components designed for logo embedding.

\subsection{Model Architecture}
\begin{figure}[h]
    \centering
    \includegraphics[width = 0.8\linewidth]{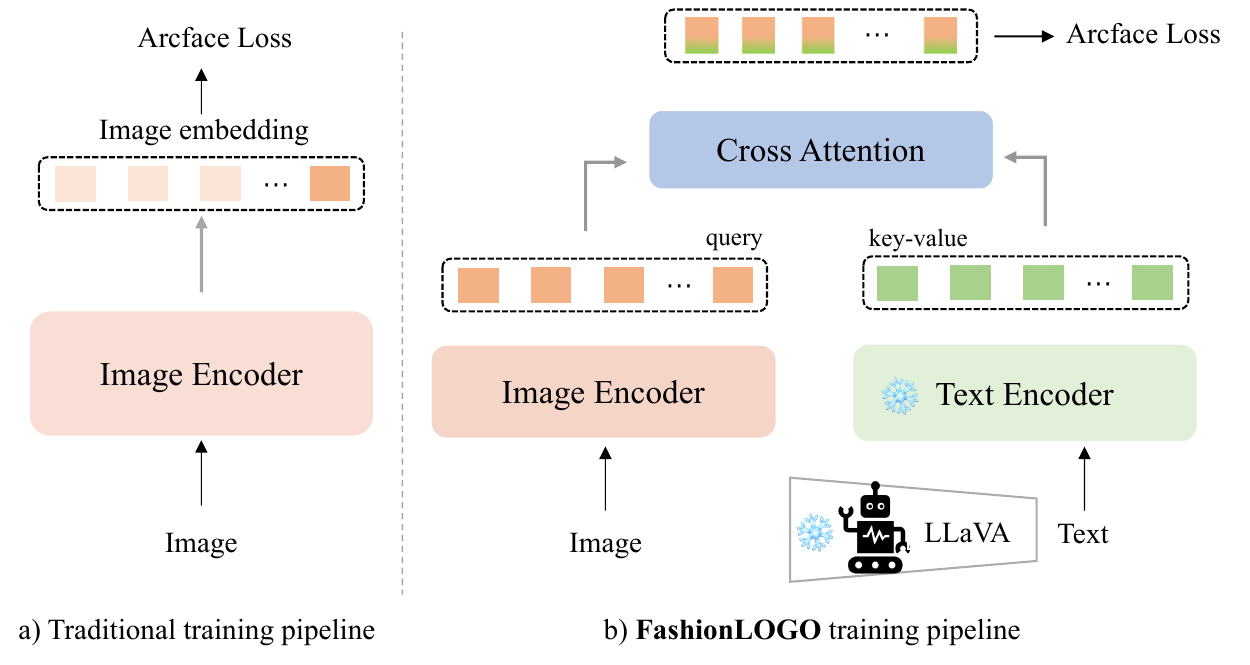}
    \caption{
    FashionLOGO framework overview.
    a) shows the traditional pipeline for training a logo embedding model and b) shows the training pipeline of FashionLOGO. We utilize the image and text encoder from CLIP to extract the image and text embeddings in which the text inputs are generated by LLaVA offline. Then, a cross-attention transformer is adopted to enhance the image embedding by learning textual embedding through the training process.
    }
    \label{fig:method}
\end{figure}

FashionLOGO is composed of three primary modules: 
(1) A visual encoder $E_v$ that extracts visual features $\mathbf{v} = E_v(x)$ from input images $x$. 
(2) A textual encoder $E_t$ that extracts text features $\mathbf{t} = E_t(y)$ from textual information $y$ generated by LLaVA~\cite{DBLP:journals/corr/abs-2304-08485}. We generate three types of text information for all images in the training set using the following prompts:
(i) \emph{\textbf{OCR prompt}}: ``What are the words in the image?" This prompt focuses on OCR and queries the text present in the image.
(ii) \emph{\textbf{Brief description prompt}}: ``Describe the image concisely." This prompt provides a simple understanding of the image.
(iii) \emph{\textbf{Detail description prompt}}: ``Describe the following image in detail." This prompt offers a detailed description of the image content. 
(3) A Representation Enhancement Module, denoted as $M$, effectively merges visual features and textual features to project them into a cohesive latent space $\mathcal{Z}$, yielding the fused feature vector $\mathbf{f} = M(\mathbf{v}, \mathbf{t}).$ We use CLIP as our image encoder and text encoder, the training and inference framework is illustrated in Figure~\ref{fig:method}.

\vspace{-10pt}

In the implementation of FashionLOGO, we use the brief description as text input. Compared to other text inputs, it has the best performance on all benchmarks. We discuss the impact of different texts on the logo embedding model in the section~\ref{result}.

\subsection{Representation Enhancement Module}
Multimodal feature fusion plays a critical role in enhancing the quality of feature representations by establishing meaningful connections between visual and textual modalities. We study three different multimodal fusion methods: MLP layer, self-attention transformer, and cross-attention transformer, as part of our representation enhancement module.

The MLP layer method involves directly concatenating visual and textual features $\mathbf{v}$ and $\mathbf{t}$ and feeding them into a lightweight multilayer perceptron (MLP) $\phi$. This method updates the weights $W$ without any conditions and has limited performance improvement.
\begin{align*}
M_{\rm MLP} &= \phi([\mathbf{v}; \mathbf{t}];W).
\end{align*}

To overcome these limitations, we introduce self-attention layers $\mathbf{S}$ between the inputs and MLP layer. Specifically, we input the concatenated features into a multi-headed self-attention layer, allowing the model to learn to attend to information from different modalities adaptively. This method effectively improves the quality of the fused feature representation.
\begin{align*}
M_{\rm SA} &= \mathbf{S}([\mathbf{v};\mathbf{t}];W_k,W_q,W_v),
\end{align*}
where $W_k, W_q,$ and $W_v$ are weight matrices.

The cross-attention transformer is another multimodal feature fusion method that utilizes cross-attention layers $\mathbf{C}$. This method enables the model to learn inter-modal dependencies between visual and textual features. In this method, visual queries $q_t$ guide the model to selectively attend to textual features $k_t$ and $v_t$, allowing the model to fully utilize the textual signals to enhance visual understanding. This selective fusion mechanism adaptively combines complementary information from both modalities. 
\begin{align*}
M_{\rm CA} &= \mathbf{C}(q_v,k_t,v_t;W_k,W_q,W_v).
\end{align*}
Overall, the fusion module consists of a total of m cross-attention layers. 
$M_{CA_{m}}(\cdot)$ aggregates information from $v_t$ to $q_v^{(m-1)}$ with multi-head cross attention and $\parallel$ denotes concatenation. In detail, $q_v^{(m-1)}$ and $v_t$ are fused by $MCA_m(q_v^{(m-1)}, v_t)=(CA_m^{(1)}\parallel,...,\parallel CA_m^{(n_h)})W^{(m)}$, where $CA_m^{(i)}$ is the $i^{th}$ cross-attention output, $n_h$ is the number of cross-attention heads and $W^{(m)}$ is a $\mathbb{R}^{C\times C}$ projection matrix. 
\begin{equation}
\begin{split}
    q^{(m)} &= q_v^{(m-1)}W_q^{(m)}, \\
    k^{(m)} &= k_tW_k^{(m)}, \\
    v^{(m)} &= v_tW_v^{(m)}, \\
    CA_{m} &= softmax(\frac{q^{(m)}{k^{(m)}}^T}{\sqrt{C/n_h}})v^{(m)},
\end{split}
\end{equation}
where $W_q^{(m)}, W_k^{(m)}, W_v^{(m)} \in \mathbb{R}^{C\times C/n_h}$ is query, key and value projection matrix, respectively.

\subsection{Training}
We train the FashionLOGO using an end-to-end approach that optimizes the similarity of feature embeddings between positive pairs and the dissimilarity between negative pairs. This is achieved using the Arcface loss.

\textbf{ArcFace loss} extends the softmax loss by introducing an additive angular margin penalty that encourages larger angular distances between classes. The goal is to maximize the cosine similarity between feature vectors of the same class while minimizing the cosine similarity between feature vectors of different classes. The loss is computed using Eq.~\eqref{eq:arc}:
\begin{equation}
\small
\mathcal{L}_{arc} = -\frac{1}{N}\sum_{i=1}^{N}\log\frac{\exp(s\cos(\theta_{y_i}+m))}{\exp(s\cos(\theta_{y_i}+m))+\sum \limits_{j\neq y_i}\exp(s\cos(\theta_{j}))}.
\label{eq:arc}
\end{equation}where $\cos(\theta_{y_i})$ is the cosine similarity between the $i$-th feature vector and its corresponding class center, $\cos(\theta_{j})$ is the cosine similarity between the $i$-th feature vector and the $j$-th class center, $m$ is the angular range, and $s$ is the feature scale parameter.

For different images of the same logo with different contents, the vector representations obtained by the model are not always similar. When describing these logo images, we will use some identical or similar sentences, and the distance between the features corresponding to this part of the text is similar, which can help to bring the gap between the image features of the same logo closer and thus get a much better image representation. 

\subsection{Inference}

During training, we add text to improve the logo representation. For inference, We discard the text encoder $E_t$ and the representation enhancement module $M$. The image encoder $E_v$ is used alone to extract the logo representation. In this way, a lightweight inference framework consists only of image encoders.
For the retrieval task, we compute the embeddings for the query logo image $\mathbf{v}_q$ and the gallery images $\mathbf{v}_\mathcal{G}$, where $\mathcal{G}=\{g_i\}_{i=1}^N$. We then calculate the similarity between the query and gallery embeddings using the cosine similarity and output the top-k most similar images as results.

\begin{table*}[htbp!]
\centering
\resizebox{0.90\textwidth}{!}{
\small
\begin{tabular}{lllllllllll} \toprule
            & \multicolumn{4}{c}{Open Brands}      & \multicolumn{3}{c}{Logo-2K+} & \multicolumn{3}{c}{IPRLogo} \\
            \cmidrule(lr){2-5}\cmidrule(lr){6-8}\cmidrule(lr){9-11}
            & Recall@1 & Recall@5 & NDCG@5 & MAP@100 & Recall@5  & NDCG@5  & MAP@100  & Recall@5  & NDCG@5  & MAP@100 \\ \midrule
PP-ShiTu    & 87.82    & 91.94    & 89.77  & 81.85 & 58.51     & 53.06   & 39.21  & 81.91     & 73.09   & 49.05 \\
ResNet-50$\star$   & 90.58    & 94.27    & 92.32  & 84.88 & 55.14     & 59.72   & 48.84  & 87.19     & 78.55   & 52.79 \\
ViT$\star$         & \underline{97.66}    & \underline{98.07}    & \underline{97.88}  & \underline{97.37} & \underline{73.00}     & \underline{67.47}   & \underline{53.54}  & \underline{85.93}     & \underline{83.87}   & \underline{71.05} \\
FashionLogo & \textbf{98.55}    & \textbf{99.07}    & \textbf{98.81}  & \textbf{98.12} & \textbf{76.36}     & \textbf{70.97}   & \textbf{57.26}  & \textbf{90.21}     & \textbf{87.80}   & \textbf{75.19} \\ \bottomrule
\end{tabular}
}
\caption{Performance comparison between other baselines and FashionLOGO on three datasets. $\star$ indicates the version of our implementation.}
\label{tab:out_of_domain}
\end{table*}

\section{Experiments Setting}

\subsection{Datasets}
We train models on Logodet3K~\cite{wang2020logodet3k} and Open Brands~\cite{jin2020open}. 

\textbf{Logodet3K~\cite{wang2020logodet3k}} is the most extensive fully annotated logo detection dataset available with 3,000 logo categories and around 200,000 manually annotated logo objects across 158,652 images. 

\textbf{Open Brands\cite{jin2020open}} contains 1,437,812 images with brands and 50,000 images without any brand. 

The two datasets contain different logo categories, we clean and mix them. As a result, we obtain a training set of about 300,000 samples covering 2,521 categories.

During the evaluation phase, more datasets were introduced to measure the in-domain and out-of-domain performance of different models. The in-domain test dataset is sampled from Open Brands~\cite{jin2020open}, excluding data used for training. We limit the number of samples per category between 20 and 50, resulting in 371 categories and 18,419 samples. As for the evaluation of the out-of-domain generalization ability, we chose the following two datasets.

\textbf{Logo-2K+\cite{wang2020logo}} contains a diverse range of logo classes from real-world logo images. It contains 167,140 images with 10 root categories and 2,341 leaf categories.

\textbf{IPRLogo} is collected from online websites by ourselves. It contains 7865 logo images with 45 categories.
\subsection{Baselines \& Evaluation }
We compare FashionLOGO against the following baselines to demonstrate its effectiveness.
\begin{itemize}[leftmargin=2em]
    \item \textbf{ResNet50~\cite{he2015deep}} is a variant of the Residual Network architecture, this design significantly improves performance on a wide range of image recognition tasks. 
    \item \textbf{ViT~\cite{dosovitskiy2021image}} is a plain, non-hierarchical architecture that enables the model to learn spatial hierarchies and features in an efficient and scalable manner, demonstrating remarkable performance on various computer vision tasks.
    \item \textbf{PP-ShiTu~\cite{wei2022ppshitu}} is an advanced image recognition and retrieval system designed to accurately identify and classify a wide range of images from various categories. 
\end{itemize}
The logos retrieved from different models are compared, for in-domain benchmarks, we report the Recall at 1 and 5 (Recall@1 and Recall@5), Normalized Discounted Cumulative Gain at 5 (NDCG@5), and Mean Average Precision at 100 (MAP@100). For out-of-domain benchmarks, we report the Recall at 5 (Recall@5), Normalized Discounted Cumulative Gain at 5 (NDCG@5), and Mean Average Precision at 100 (MAP@100). 

\section{Result and Discussion} \label{result}

We compare FashionLogo with other baselines on a variety of widely recognized benchmarks. The results show that FashionLogo consistently improves the performance of retrieval, both in the domain and out of the domain. Compared to the best baseline ViT, FashionLogo improves by an average of 1\% across all metrics on the in-domain dataset Open Brands, and by 3\% and 4\% on the out-of-domain datasets Logo-2K+ and IPRLogo, respectively. This suggests that the use of text in the training phase can optimize the embedding of the product image to focus on the logo information. This ability can be generalized to a wide range of scenarios, not just limited to training-related data.
\begin{table}[htbp!]
\small
\resizebox{0.8\linewidth}{!}{
\begin{tabular}{llll} \toprule
Text input & Recall@5 & NDCG@5 & MAP@100 \\ \midrule
w Brief & \textbf{90.21}    & \textbf{87.80}  & \textbf{71.05} \\
w OCR+Brief & 89.56    & 86.44  & 69.82 \\
w OCR+Detail  & 89.43    & 85.89  & 70.00 \\
w Detail   & 88.92    & 86.21  & 69.20 \\
w/o\,Text   & 87.40    & 84.50  & 65.58 \\ \bottomrule
\end{tabular}
}
\caption{Influence of different text inputs on FashionLOGO.}
\label{tab:results-with_text}
\end{table}
\vspace{-10pt}

To analyze the improvement of the model to represent logos with the introduction of different types of text, we vary the input text and observe the corresponding change in the performance of the model on the IPRLogo. The results are shown in the Table~\ref{tab:results-with_text}. It is not a fact that the more detailed the text, the more helpful the visual model will be in extracting the features of the logo. Neither the OCR text, the detailed description text, nor a combination of both, exceed the impact of the brief text. A brief description highlights the core of the content displayed in the product images, which is similar to the purpose of a logo design. This means that FashionLogo is still effective for logos that don't include text, and it makes the visual embedding focus on the core area of the product images.

The introduction of text has a stable improvement in the representation of the logo compared to the use of visual models alone. We design a special fusion module to maximize the use of text input. To illustrate the importance of fusion methods, we compare the performance of FashionLOGO on IPRLogo when using different text fusion methods. As can be seen from the results presented in Table~\ref{tab:fusion-performance}, the fusion method of $Cross\, Attn$ can utilize the text input best to improve the logo embedding compared to the $MLP$ and $Self\, Attn$.

\begin{table}[htbp!]
\centering
\small
\resizebox{0.8\linewidth}{!}{
\begin{tabular}{llllll}
\toprule

Fusion method & Recall@5 & NDCG@5 & MAP@100 \\ \midrule
$Cross\,Attn$ & \textbf{90.21} & \textbf{87.80} & \textbf{71.05} \\
$MLP$       & 87.81 & 85.92 & 70.39 \\
$Self\,Attn$ & 89.17 & 85.90 & 68.72 \\\bottomrule
\end{tabular}
}
\caption{Performance comparison between different modal fusion methods.}
\label{tab:fusion-performance}
\end{table}
\vspace{-10pt}

\begin{figure}[h]
    \centering
    \includegraphics[width=0.8\linewidth]{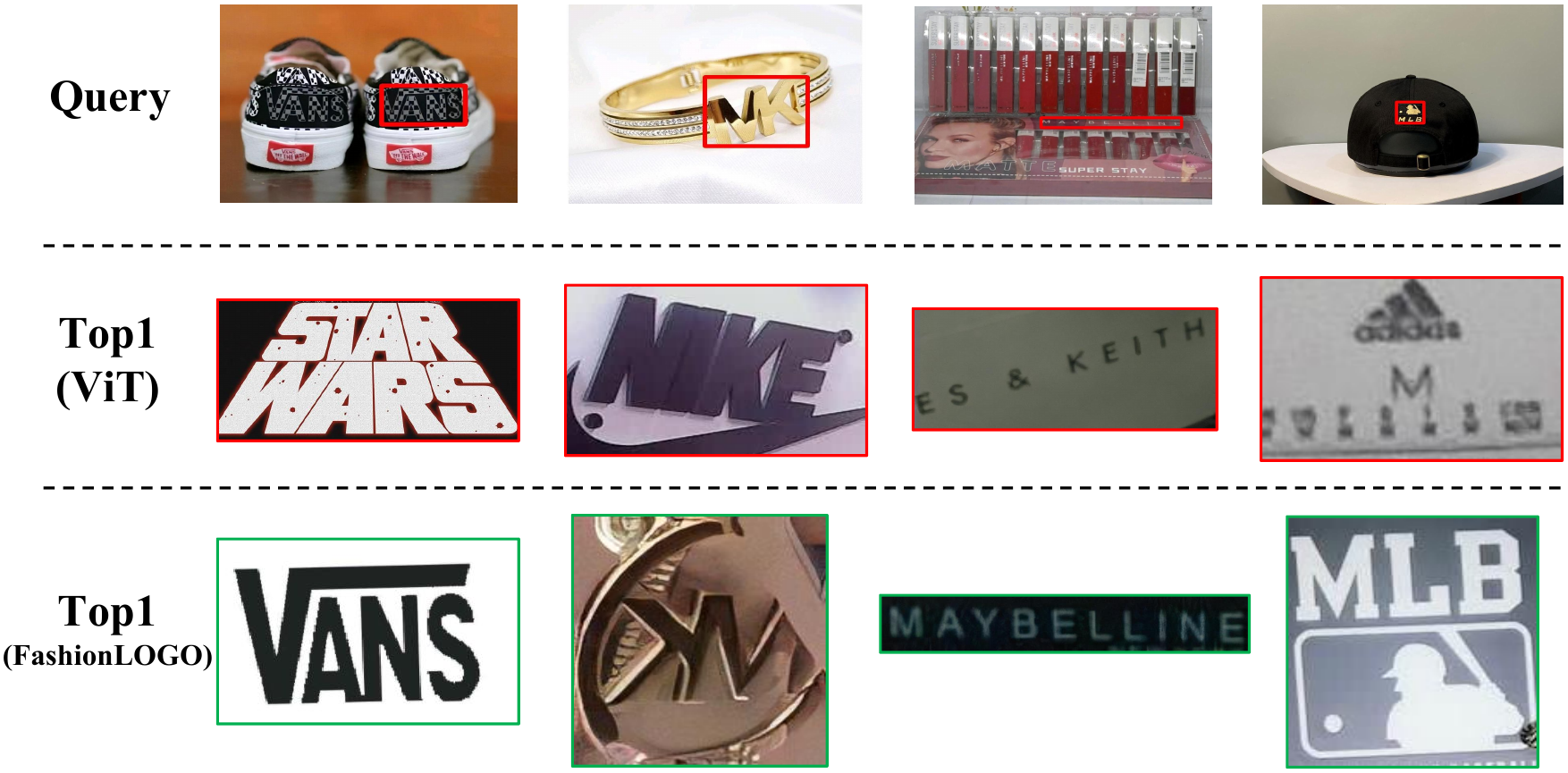}
    \caption{
    Qualitative analysis. The three rows represent queries, and the Top-1 results are obtained from ViT and FashionLOGO, respectively. The correct results are in green frame and the incorrect ones are in red.
    }
    \label{fig:hardcase}
\end{figure}

To visually illustrate FashionLOGO's performance enhancements, we analyze cases to observe the performance improvement of FashionLogo compared to ViT. We extracted some logo images to retrieve similar logos from the remaining logos of IPRLogo. Figure \ref{fig:hardcase} shows the Top-1 results of ViT and FashionLOGO.
We find that the ViT struggles to distinguish the \textit{"VANS"} logo from \textit{"STAR WARS"} logo, and the \textit{"MK"} logo from \textit{"NIKE'}. The reason is these logos share similar textual contents which makes them difficult to distinguish. On the other hand, FashionLOGO can successfully identify the correct logos.This shows that the incorporation of auxiliary text information from LLMs has empowered FashionLOGO to acquire more robust visual features, particularly when dealing with logos containing textual elements.

\vspace{-10pt}

\section{Conclusion}

This study introduces FashionLOGO, which improves the visual model's ability to generate logo embedding utilizing auxiliary text generated by the Multimodal Large Language Model (MLLM). Rigorous evaluations demonstrate FashionLOGO's superior performance in fashion image retrieval in applications like e-commerce. This suggests that text can be used to optimize the embedding of logos in product images by highlighting the area of the logo in the image. This improvement is stable and robust enough to be successfully generalized to zero-shot scenarios.
\bibliographystyle{ACM-Reference-Format}
\bibliography{sample-base}

\end{document}